%% file: main.tex
\documentclass[runningheads]{llncs}
\usepackage[T1]{fontenc}
\usepackage{graphicx,verbatim}
\usepackage{amssymb}
\usepackage{amsmath}
\usepackage{booktabs}
\usepackage{xcolor} 
\usepackage{hyperref} 
\usepackage{wrapfig}
\usepackage{caption}

\hypersetup{
    colorlinks=true, 
    linkcolor=red,      
    citecolor=blue,   
    filecolor=magenta, 
    urlcolor=magenta       
}
\usepackage{multirow}
\usepackage{makecell}
\usepackage{subcaption}

\makeatletter
\renewcommand*{\@fnsymbol}[1]{\ifcase#1\or$\ast$\else\@$\dagger$\fi}
\makeatother

\begin{document}

\title{PDC-Net: Pattern Divide-and-Conquer Network for Pelvic Radiation Injury Segmentation}

\titlerunning{Pelvic Radiation Injury Segmentation}
\author{Xinyu Xiong\inst{1}\thanks{Equal contribution.}
\and Wuteng Cao\inst{2}$^*$
\and Zihuang Wu\inst{3}$^*$
\and Lei Zhang\inst{1}  
\and Chong Gao\inst{1}
\and Guanbin Li\inst{1}$^\dagger$
\and Qiyuan Qin\inst{4,5}\thanks{Corresponding authors.}
}
\authorrunning{X. Xiong et al.}
\institute{
$^1$School of Computer Science and Engineering, Sun Yat-sen University\\ 
$^2$Department of Radiology, The Sixth Affiliated Hospital, Sun Yat-sen University\\ 
$^3$School of Computer and Information Engineering, Jiangxi Normal University\\
$^4$Department of General Surgery, The Sixth Affiliated Hospital, Sun Yat-sen University\\
$^5$Guangdong Provincial Key Laboratory of Colorectal and Pelvic Floor Diseases, Biomedical Innovation Center, The Sixth Affiliated Hospital, Sun Yat-sen University\\
\email{liguanbin@mail.sysu.edu.cn, qinqy3@mail.sysu.edu.cn}}

\maketitle 
\begin{abstract}
Accurate segmentation of Pelvic Radiation Injury (PRI) from Magnetic Resonance Images (MRI) is crucial for more precise prognosis assessment and the development of personalized treatment plans. However, automated segmentation remains challenging due to factors such as complex organ morphologies and confusing context. To address these challenges, we propose a novel Pattern Divide-and-Conquer Network (PDC-Net) for PRI segmentation. The core idea is to use different network modules to “divide” various local and global patterns and, through flexible feature selection, to “conquer” the Regions of Interest (ROI) during the decoding phase. Specifically, considering that our ROI often manifests as strip-like or circular-like structures in MR slices, we introduce a Multi-Direction Aggregation (MDA) module. This module enhances the model's ability to fit the shape of the organ by applying strip convolutions in four distinct directions. Additionally, to mitigate the challenge of confusing context, we propose a Memory-Guided Context (MGC) module. This module explicitly maintains a memory parameter to track cross-image patterns at the dataset level, thereby enhancing the distinction between global patterns associated with the positive and negative classes. Finally, we design an Adaptive Fusion Decoder (AFD) that dynamically selects features from different patterns based on the Mixture-of-Experts (MoE) framework, ultimately generating the final segmentation results. We evaluate our method on the first large-scale pelvic radiation injury dataset, and the results demonstrate the superiority of our PDC-Net over existing approaches.


\end{abstract}

\section{Introduction}
In 2020, the global incidence of pelvic malignancies exceeded 4 million~\cite{GLOBALCAN20,Interpretation}, representing a significant and escalating threat to public health worldwide. Pelvic radiotherapy~\cite{CA2016} plays a crucial role in the tumor local control, systemic cure, and palliation for patient survival, making it an indispensable therapeutic approach. However, radiation exposure to normal tissues also brings damage to the pelvic and abdominal organs. Pelvic Radiation Injury (PRI) may jeopardize the physical health and quality-of-life of patients. When the precise diagnosis of PRI is difficult as for varied symptoms and nonspecific signs, Magnetic Resonance Imaging (MRI) could provide details for repeated evaluation of pelvic structure and surrounding tissues. For radiologists, PRI diagnosis is based on typical features as diffuse edema or fibrosis within irradiated fields. Further quantitative assessment requires segmentation and delineation of Regions of Interest (ROI), a process that often costs huge labor and time. In addition, manual operations are prone to errors of judgments, such as missed or redundant segmentation, which limits the reliability. Therefore, an automated algorithm to identify PRI regions from MRI could effectively assist clinicians in comprehensive diagnosis, progression monitoring, and treatment planning.

Although medical image segmentation have made promising progress in recent years~\cite{SAM2UNet,MICCAI15_UNet,CVPR24_EMCAD}, the design of methods to accurately identify PRI regions from MRI remains an unsolved challenge. Directly applying existing general methods does not yield optimal results, as PRI segmentation presents unique challenges that require tailored solutions. Specifically, our goal is to identify the rectal intrinsic muscle layer, anal sphincter muscle, puborectalis muscle, and full bladder layer in MR slices, while determining whether these regions are injured. Injuries are marked as positive pixels, while non-injured regions are defined as negative. As illustrated in Fig.~\ref{fig:title}, this task is challenging for two key reasons. First, unlike existing works where the ROIs typically consist of solid organs or lesions, our ROIs involve hollow organs and muscles, which typically exhibit circular (e.g., bladder, rectum) or strip-like (e.g., muscles) shapes, making the use of traditional square convolutions less effective for accurate feature matching. Second, in the same slice, both positive and negative regions may be present simultaneously, increasing the potential for class confusion and further increases the difficulty of accurate segmentation.

To address the challenges outlined above, we propose a novel Pattern Divide-and-Conquer Network (PDC-Net) in this paper. The network incorporates several carefully designed components to accurately identify PRI regions. First, we introduce a Multi-Direction Aggregation (MDA) module, which applies strip convolutions in four directions. Compared to traditional square convolutions, this approach better captures the shape of the target organ. Additionally, we propose a Memory-Guided Context (MGC) module that maintains dataset-level states for cross-image modeling. This enables the network to effectively distinguish complex semantics. Lastly, we present an Adaptive Fusion Decoder (AFD) that dynamically selects features from different patterns, resulting in better segmentation outputs. Overall, our contributions can be summarized as follows: (1) To the best of our knowledge, we are the first to design an AI-based method for the automatic recognition of PRI from MRI, providing an important reference for future research in this area. (2) To address challenges such as complex morphology and confusing context, we have developed three novel modules: MDA, MGC, and AFD, each carefully crafted to enhance segmentation accuracy. (3) To validate the superiority of our proposed PDC-Net, we conducted experiments on a large in-house dataset, demonstrating that our approach achieves state-of-the-art performance. 

\input{figures/fig_title}

\input{figures/fig_framework}

\section{Method}

\subsection{Multi-Direction Aggregation Module}
\label{sec:MDA}
 
In many current medical segmentation methods~\cite{MICCAI15_UNet,TMI19_UNet++,CVPR24_EMCAD}, convolutions typically use standard square kernels to extract feature maps within square windows. This approach is effective for segmenting solid organs or lesions with relatively regular shapes. However, for injury segmentation, the primary focus is on hollow organs and muscles, whose surfaces exhibit circular or strip-like shapes. In such cases, existing square convolution operations may inadvertently capture more irrelevant background information from neighboring pixels, leading to inefficiencies.

To address the aforementioned challenge, we design the Multi-Direction Aggregation (MDA) module, which focuses on refining the features output by the backbone to capture complex target shapes. While some existing methods~\cite{NIPS22_SegNeXt} also use strip convolutions for image segmentation, their approach only incorporates horizontal and vertical convolutions. This works well for targets with more rectangular shapes, such as the bladder. However, for other targets, like muscles, which appear diagonal in the slice, vertical or horizontal convolutions become less efficient. To address this issue, we further extend the strip convolution to include diagonal directions, enabling the model to better capture the irregular shapes of various targets.

Specifically, given the encoder feature $f_{i}$, we first use a 1×1 depthwise convolution to adjust feature. Then we divide it into four parts along the channel, and obtain $f_i^{P1}, f_i^{P2}, f_i^{P3}, f_i^{P4}$. The four parts are fed into strip convolutions with horizontal ($\text{Conv}_{9\time9}^{\rightarrow}$), vertical ($\text{Conv}_{9\time9}^{\downarrow}$), left diagonal ($\text{Conv}_{9\time9}^{\searrow}$), and right diagonal ($\text{Conv}_{9\time9}^{\nearrow}$) directions and kernel size 9 to capture features of different patterns. The captured features are concatenated along the channel to return to their original shape to obtain the aggregated feature. In addition, for the input feature $f_i$, we also feed it into an auxiliary max pooling branch, to obain the detail-enhanced feature. The aggregated feature and enhanced feature are multiply in element-wise and sent to 1×1 convolution to obtain the final output of the MPA module. The above operations can be represented as follows: 
\begin{eqnarray}
f_{i}^{Pj} & = & \text{Conv}_{1\times9}[x,y](\text{Split}(\text{DWConv}_{1 \times 1}(f_{i}))),j\in\{1,2,3,4\},\\
f_{mda} & = & \text{Conv}_{1\times 1}(\text{ReLU}(\text{BN}{(\text{Cat}[f_{i}^{P1},f_{i}^{P2},f_{i}^{P3},f_{i}^{P4}])})\otimes \text{MaxPool}_{3 \times 3}(f_i))
\end{eqnarray}
where \([x, y]\) denotes the direction vector of the strip convolution, where \([1, 0]\), \([0, 1]\), \([1, 1]\), and \([-1, 1]\) correspond to horizontal, vertical, left diagonal, and right diagonal convolutions, respectively.
\subsection{Memory-Guided Context Module}
\label{sec:MGC}
The MDA module described above is focus on capturing local patterns such as organ shape. To get a more discriminating global context, such as weather a region is positive or negative, further mining the very deep features is needed. Existing practices usually focus on obtain a larger receptive field by stacking different dilated convolution blocks or using self-attention operations~\cite{ECCV24_EDAFormer}. However, these methods do not take into account the effects of class imbalance. Due to this severe imbalance, these methods are more susceptible to interference from a large amount of background information, thereby reducing their ability to distinguish between the semantics of negative and positive classes.

To better capture global semantics and reduce the interference of irrelevant background, we propose a Memory-Guided Context (MGC) module. Memory modules are often used in video segmentation~\cite{ICCV19_Memory} to capture the context of time dimensions. This time dimension can also be viewed as the dataset dimension, and we are inspired by this to use a memory to enable the network to selectively and progressively accumulate knowledge related to the foreground class.

To be specific, given the input features $f_{in}\in\mathbb{R}^{H\times W\times M}$, we use a sliding window of size $N$ to slice them without overlapping. In this way, we can obtain both background patches and foreground patches, and enhance these foreground patches in the next. We then apply max pooling and average pooling to each patch $f_{patch}\in\mathbb{R}^{\frac{H}{N} \times \frac{W}{N} \times M}$ separately to extract key features. Then, we use fully connected (FC) layer on the max pooled and average pooled features for dimensionality reduction, eliminating extraneous background slices:
\begin{eqnarray}
f_{patch}^{'} & = & \text{FC}(\text{Cat}[\text{Avg}(f_{patch}),\text{Max}(f_{patch})]),
\end{eqnarray}
Additionally, we employ a dynamic weighting approach on the max pooled and average pooled features to model the regions of interest:
\begin{eqnarray}
f_{patch}^{w} & = & f_{patch}^{'}\otimes Avg(f_{patch}) \oplus  (1-f_{patch}^{'})\otimes Max(f_{patch})
\end{eqnarray}

Subsequently, we designed a memory bank \(\mathcal{M} \in \mathbb{R}^{1\times 1\times S \times K}\) to accumulate dataset-level target representations, where \(K\) denotes the capacity of the memory bank. We multiply \(f_{patch}^{w}\in \mathbb{R}^{1\times 1\times S}\) with \(\mathcal{M}\) to obtain \(f_{MB} \in \mathbb{R}^{1\times 1\times S \times K}\) and use the softmax function to generate a similarity coefficient matrix \(\mathcal{S} \in \mathbb{R}^{1 \times 1 \times S}\), which measures the similarity between the input $f_{patch}^{w}$ and elements in \(\mathcal{M}\). We multiply \(\mathcal{S}\) with \(f_{MB}\) to perform a weighted sum of the input features based on their relevance, resulting in the organ-related weights \(\hat{f}_{patch}^{w}\). Next, we use a FC layer to upscale \(\hat{f}_{patch}^{w}\) to the dimension of \(f_{patch}\), and perform an element-wise multiplication with \(f_{patch}\) to obtain background-independent patches. Finally, the slices are restored to the original feature map through the reverse sliding window operation.

\subsection{Adaptive Fusion Decoder}
\label{sec:AFD}
Many existing decoders in medical segmentation networks adhere to the U-Net~\cite{MICCAI15_UNet} style, progressively recovering information through a single bottom-up pathway. However, this structure implicitly demands that the same decoder features address multiple segmentation challenges simultaneously, such as unclear boundaries or small targets. This constraint increases the learning difficulty of the network. Recently, the Mixture-of-Experts (MoE)~\cite{MoE,MMAsia24_MoEPolyp} architecture has garnered great attention in the field of natural language processing. Its core concept involves constructing a group of experts, each of which excels at processing distinct patterns. By dynamically adjusting the weights of these experts, the architecture can flexibly handle various samples, thereby enhancing the overall performance of the network. However, the vanilla MoE decoder lacks feature flow between different stages, resulting in insufficient expert representation. Therefore, we propose an Adaptive Fusion Decoder (AFD) based on the MoE architecture, which encourages interaction among different expert features to enhance performance in the PRI segmentation task.

Specifically, our decoder receives the features of previous phases as input. We first divide each feature into four equal parts along the channel dimension. Then, we use patch shuffle to select each piece of each feature and rejoin them to obtain the mixed feature. This process effectively promotes the interaction of features in different stages:
\begin{eqnarray}
f^{i}_{ps} & = & PS(f_{MDA}^{1},f_{MDA}^{2},f_{MDA}^{3},f_{MDA}^{4}),i\in\left \{1,2,3,4  \right \},
\end{eqnarray}
Next, we apply depthwise convolution with kernel sizes of \((1 \times n, n \times 1)\) to achieve expert customization, where \(n \in \{3, 5, 7, 9\}\), to the features from these four different stages to tailor the capabilities of different experts. We then concatenate these expert features along the channel dimension and use a \(1 \times 1\) depthwise convolution to further eliminate redundant features, resulting in \(F_{cat}\):
\begin{eqnarray}
\hat{f} & = & \text{DWConv}_{1\times 1}(\text{Cat}[\text{DWConv}_{n\times 1}(\text{DWConv}_{1\times n}(f^{i}_{ps})))]),n  =  \{3,5,7,9\},
\end{eqnarray}
Subsequently, we introduce an adaptive expert feature fusion mechanism to generate the final output. Specifically, we apply adaptive pooling and a fully connected layer to the concatenated features \(\hat{f}\) to reduce the dimensionality to the number of experts. A sigmoid activation function is used to generate the weights for the outputs of different experts. Finally, we obtain the prediction result of each expert through a \(1 \times 1\) depthwise convolution and perform element-wise multiplication of the expert weights with the expert prediction results to achieve the final output.
\begin{eqnarray}
f_{out}& = &  \text{DWConv}_{1\times 1}(\hat{f})\otimes  \sigma (\text{FC}(\text{AvgPool}(\hat{f}))
\end{eqnarray}

\section{Experiments}

\subsection{Experimental Setup}
\subsubsection{Datasets.}
Due to the absence of a publicly available dataset for pelvic radiation injury segmentation, we primarily conducted experiments using our in-house dataset, which was obtained from the Sixth Affiliated Hospital, Sun Yat-sen University. This study was carried out following a protocol approved by the Ethics Committee of the Sixth Affiliated Hospital, Sun Yat-sen University. The dataset includes T2-weighted MRI sequences, where radiologists manually delineated the rectal intrinsic muscle layer, anal sphincter muscle, puborectalis muscle, and full bladder layer on each axial image slice. These four regions are annotated as injured (positive) or not (negative). The final dataset comprises a total of 344 pelvic cases (8049 images), with 274 cases (6413 images) allocated for training and the remaining 70 cases (1636 images) reserved for testing.

\input{table_sota}

\subsubsection{Implementation Details.}
The framework is implemented using the PyTorch library, with all experiments conducted on an NVIDIA 3090 GPU. We utilize the CosineAnnealingLR schedule to adjust the learning rate, starting at \(1 \times 10^{-4}\) and gradually decreasing to a minimum of \(1 \times 10^{-6}\). Training is performed over 100 epochs using the AdamW optimizer with a weight decay of \(5 \times 10^{-4}\). To optimize the network, we employ both cross-entropy loss and Dice loss. Input images are uniformly resized to \(512 \times 512\), with a batch size of 16.

\subsubsection{Compared Methods.}
We compared our PDC-Net with the following methods: U-Net~\cite{MICCAI15_UNet}, UNet++~\cite{TMI19_UNet++}, TransUNet~\cite{MedIA24_TransUNet}, H2Former~\cite{TMI23_H2Former}, UNETR++~\cite{TMI24_UNETR++}, and EMCAD~\cite{CVPR24_EMCAD}. For quantitative evaluation, we employed four metrics: the Dice Similarity Coefficient (DSC), Matthews Correlation Coefficient (MCC), Accuracy (ACC), and Hausdorff Distance (HD) score. It is worth noting that, due to the class imbalance between foreground and background, the DSC and MCC for PRI segmentation tend to be relatively low, while the ACC is relatively high.

\subsection{Result Analysis}
\noindent
\textbf{Quantitative Comparison} results are shown in Table~\ref{tab:compare_sota}. Our PDC-Net consistently outperforms other methods. For example, in the negative class, PDC-Net improves the DSC and MCC by 4.18\% and 2.35\%, respectively, compared to the U-Net baseline. When compared to the most competitive EMCAD, DSC and MCC are improved by 2.52\% and 1.93\%, respectively. Additionally, compared to the 3D method UNETR++, our method achieves improvements of 2.96\% and 2.33\%. These experimental results demonstrate that our method is more suitable for injury segmentation and outperforms existing approaches.

\input{figures/fig_results}

\noindent
\textbf{Qualitative Comparison} results are shown in Fig.~\ref{fig:results}. In the first row of images, both positive targets (e.g., muscles) and negative targets (e.g., bladder) are present simultaneously. Our method is able to more accurately distinguish between positive and negative targets without causing class confusion. In the second row, compared to other methods, our approach produces more complete and continuous predictions without misidentifying irrelevant regions.

\input{table_abla}

\subsection{Ablation Study}
To validate the effectiveness of each component of PDC-Net, we conducted extensive ablation experiments, and the DSC results are shown in Table~\ref{tab:abla}.

\noindent
\textbf{Multi-Direction Aggregation Module.} We considered two variants of the MDA module: one that only uses point convolutions (PConv) to align channels and one that only uses vertical and horizontal patterns (Dual), without diagonal strip convolutions. As shown in Table~\ref{tab:abla_mda}, our MDA shows an improvement of 2.61\% on the neg class compared to the "PConv" mode.

\noindent
\textbf{Memory-Guided Context Module.} 
We considered using two other feature enhancement modules: Embedding-Free Attention (EFA)~\cite{ECCV24_EDAFormer} and Multi-Scale Representation (MSR)~\cite{ICLR24_VMFormer}. As shown in Table~\ref{tab:abla_mgc}, our AFD shows an improvement of 1.28\% on the neg class compared to EFA.

\noindent
\textbf{Adaptive Fusion Decoder.} 
We considered using two other decoder styles: the U-shape decoder~\cite{MICCAI15_UNet} and the vanilla MoE decoder~\cite{MoE,MICCAI22_Patcher} (without patch shuffling, expert customization, etc.). As shown in Table~\ref{tab:abla_afd}, our AFD shows an improvement of 2.88\% on the neg class compared to the U-shape decoder.

\section{Conclusion}
In this paper, we propose the first method for pelvic radiation injury segmentation. The proposed PDC-Net carefully addresses several challenges, such as complex organ morphologies and confusing context. Extensive experiments on a large in-house dataset show that our framework achieves state-of-the-art performance.

\begin{credits}
\subsubsection{Acknowledgements.} 
This work is supported in part by the National Natural Science Foundation of China under Grant NO.~62322608 and~82473561, in part by the Guangdong Basic and Applied Basic Research Foundation under Grant NO.~2024A1515010255.

\subsubsection{\discintname}
The authors declare that they have no competing interests.
\end{credits}

\bibliographystyle{splncs04}
\bibliography{refs}

\end{document}

%% file: figures/fig_title.tex
\begin{figure*}[t]
    \centering
    \includegraphics[width=0.8\linewidth]{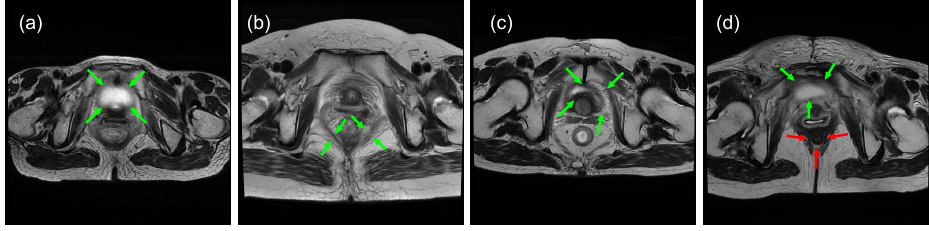}
    \caption{Challenges in recognizing the radiation injuried regions: (a) circular-like structure (b) strip-like structure (c) irregular shape (d) confusing context.
    }
    \label{fig:title}

\end{figure*}

%% file: figures/fig_framework.tex
\begin{figure*}[t]
    \centering
    \includegraphics[width=0.9\linewidth]{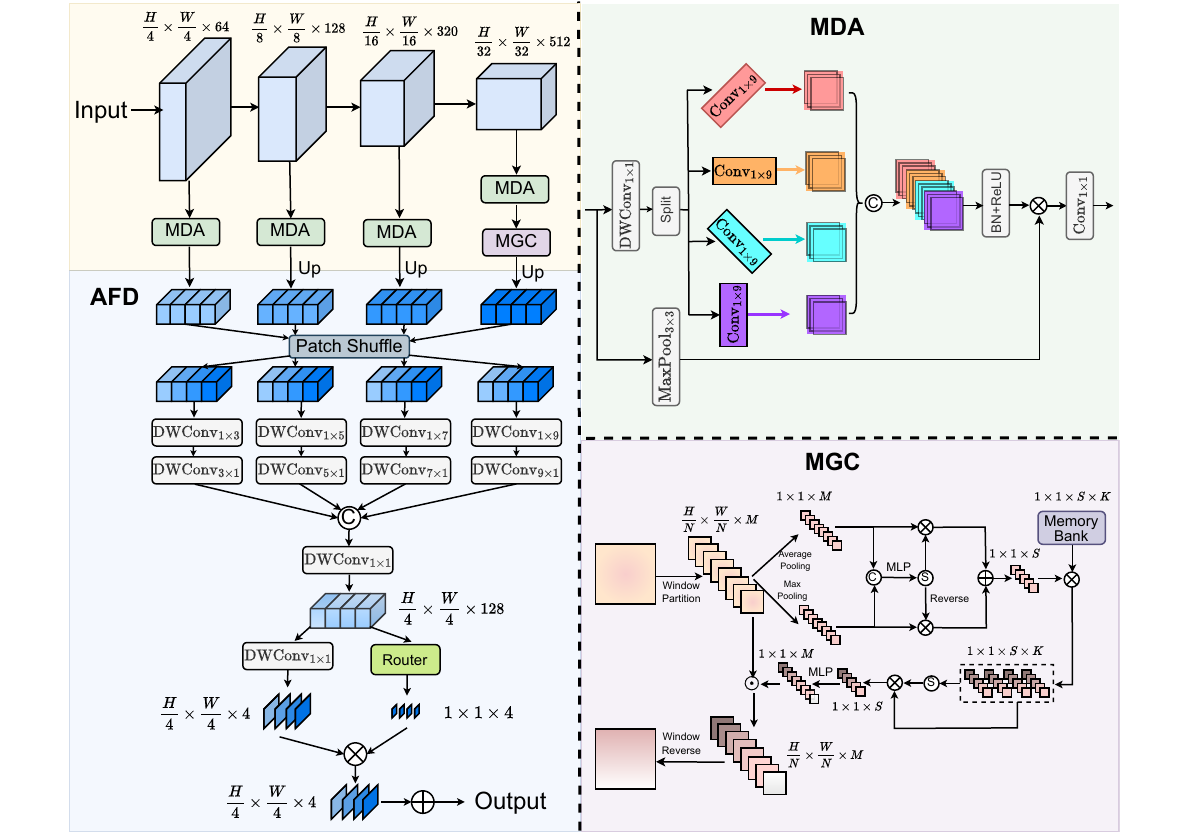}
    \caption{The overall framework of our proposed PDC-Net, which consists of four main components: the PVT-v2-b2~\cite{PVTv2} encoder, the MDA module (Sec~\ref{sec:MDA}), the MGC module (Sec~\ref{sec:MGC}), and the AFD (Sec~\ref{sec:AFD}).}
    \label{fig:framework}
\end{figure*}

%% file: table_sota.tex
\begin{table}[t]
\centering
\caption{Quantitative comparison with other medical image segmentation methods. PRI-Neg and PRI-Pos represent the results for the negative and positive classes, respectively. The best results are highlighted in bold.}
\label{tab:compare_sota}
\renewcommand\arraystretch{1.2}
\renewcommand\tabcolsep{2.0pt}
\scalebox{1.0}{
\begin{tabular}{l|cccc|cccc} 
\Xhline{1pt}
\multirow{2}{*}{Method} & \multicolumn{4}{c|}{PRI-Neg} & \multicolumn{4}{c}{PRI-Pos}     \\ 
\cline{2-9}
& DSC & MCC & ACC & HD & DSC & MCC & ACC & HD \\ 
\hline
U-Net~\cite{MICCAI15_UNet} & 44.61 & 47.96 & 99.17 & 38.15 & 39.97 & 44.13 & 98.96 & 45.12 \\
UNet++~\cite{TMI19_UNet++}  & 44.60 & 48.11 & 99.17 & 39.16 & 39.16 & 43.29 & 98.98 & 45.23 \\
TransUNet~\cite{MedIA24_TransUNet} & 41.46 & 44.88 & 99.14 & 33.08 & 41.64 & 45.79 & 98.95 & 38.87\\
H2Former~\cite{TMI23_H2Former} & 45.55 & 48.01 & 99.30 & 26.98 & 45.59 & 48.40 & 99.16 & 39.30 \\
UNETR++~\cite{TMI24_UNETR++} & 45.83 & 47.98 & 99.34 & 21.53 & 45.71 & 48.88 & 99.08 & 40.45 \\
EMCAD~\cite{CVPR24_EMCAD} & 46.27 & 48.38 & 99.32 & 21.90 & 46.30 & 49.52 & 99.12 & 34.58\\

\hline
\textbf{Ours}  & \textbf{48.79} & \textbf{50.31} & \textbf{99.38} & \textbf{21.01} & \textbf{49.12} & \textbf{50.98} & \textbf{99.26} & \textbf{33.14} 
\\
\Xhline{1pt}
\end{tabular}
}
\end{table}

%% file: figures/fig_results.tex
\begin{figure*}[t]
    \centering
    \includegraphics[width=0.8\linewidth]{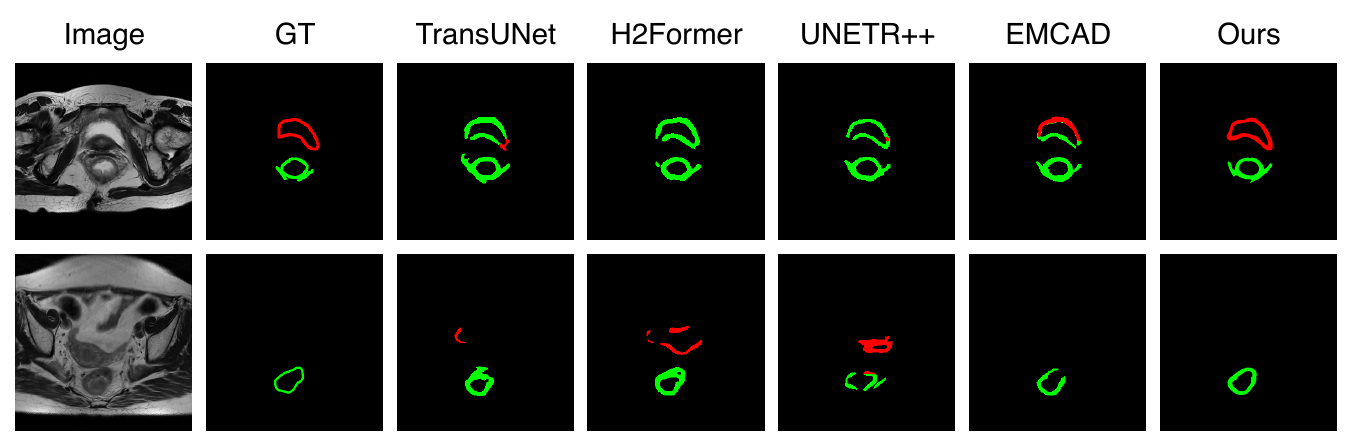}
    \caption{PRI segmentation results of different methods, where red pixels represent negative areas and green pixels represent positive areas.}
    \label{fig:results}
\end{figure*}

%% file: table_abla.tex
\begin{table}[t]
\caption{Ablation experiments on the proposed modules.}
    \centering
    \scalebox{1.0}{
    \begin{subtable}[t]{0.3\textwidth}
        \centering
        \caption{Results about MDA.}
        \label{tab:abla_mda}
        \begin{tabular}{l|cc}
            \toprule
            Variant & Neg & Pos \\
            \midrule
            PConv & 46.18 & 47.39 \\
            Dual & 47.23 & 48.01 \\
            \textbf{MDA} & \textbf{48.79} & \textbf{49.12} \\
            \bottomrule
        \end{tabular}
    \end{subtable}
    \hfill
    \begin{subtable}[t]{0.3\textwidth}
        \centering
        \caption{Results about MGC.}
        \label{tab:abla_mgc}
        \begin{tabular}{l|cc}
            \toprule
            Variant & Neg & Pos \\
            \midrule
            EFA~\cite{ECCV24_EDAFormer} & 47.51 & 47.68  \\
            MSR~\cite{ICLR24_VMFormer} & 46.93 & 47.27  \\
            \textbf{MGC} & \textbf{48.79} & \textbf{49.12}  \\
            \bottomrule
        \end{tabular}
    \end{subtable}
    \hfill 
    \begin{subtable}[t]{0.3\textwidth}
        \centering
        \caption{Results about AFD.}
        \label{tab:abla_afd}
        \begin{tabular}{l|cc}
            \toprule
            Variant & Neg & Pos \\
            \midrule
            U-shape~\cite{MICCAI15_UNet} & 45.91 & 44.12 \\
            MoE~\cite{MoE,MICCAI22_Patcher} & 47.33 & 47.58 \\
            \textbf{AFD} & \textbf{48.79} & \textbf{49.12} \\
            \bottomrule
        \end{tabular}
    \end{subtable}
    }
    \label{tab:abla}
\end{table}